# Efficient Convolutional Neural Networks for Diacritic Restoration

**Sawsan Alqahtani**[*,2,3] and **Ajay Mishra**[1] and **Mona Diab**[1,2]
[1]AWS, Amazon AI
[2]The George Washington University
[3]Princess Nourah Bint Abdul Rahman University
sawsanq@gwu.edu, misaja@amazon.com, diabmona@amazon.com

## Abstract

Diacritic restoration has gained importance with the growing need for machines to understand written texts. The task is typically modeled as a sequence labeling problem and currently Bidirectional Long Short Term Memory (BiLSTM) models provide state-of-the-art results. Recently, Bai et al. (2018) show the advantages of Temporal Convolutional Neural Networks (TCN) over Recurrent Neural Networks (RNN) for sequence modeling in terms of performance and computational resources. As diacritic restoration benefits from both previous as well as subsequent timesteps, we further apply and evaluate a variant of TCN, Acausal TCN (A-TCN), which incorporates context from both directions (previous and future) rather than strictly incorporating previous context as in the case of TCN. A-TCN yields significant improvement over TCN for diacritization in three different languages: Arabic, Yoruba, and Vietnamese. Furthermore, A-TCN and BiLSTM have comparable performance, making A-TCN an efficient alternative over BiLSTM since convolutions can be trained in parallel. A-TCN is significantly faster than BiLSTM at inference time (270%∼334% improvement in the amount of text diacritized per minute).

## 1 Introduction

A diacritic is a mark that is added above, below, or within a letter, constructing a new letter or characterizing it with a different sound (Wells, 2000). In languages such as Arabic and Hebrew, some vowels are not part of the alphabet and diacritics are used for vowel specification. Moreover, many languages that include diacritics such as Yoruba and Vietnamese sometimes omit them in writing for various reasons such as difficulty in typing diacritics on keyboards or digitizing electronic text (Scannell, 2011).

Arabic, on the other hand, considers diacritics as an optional choice in writing even in formal settings,[1] and familiarity with the language is relied upon to derive the meaning based on context. Although people can extrapolate missing diacritics with near perfect accuracy in such languages, missing diacritics pose a challenge for computational models due to increased ambiguity.

Diacritization is the process of automatically restoring missing diacritics for each character in written text. It is important in many NLP applications such as automatic speech recognition (Vergyri and Kirchhoff, 2004) and speech synthesis (Ungurean et al., 2008). Most state-of-the-art diacritization models use Bidirectional Long Short Term Memory (BiLSTM) (Hochreiter and Schmidhuber, 1997) as a sequential classification problem, or as a sequence-to-sequence transduction problem to convert the original text into diacritized form (Orife, 2018; Zalmout and Habash, 2017; Belinkov and Glass, 2015). Generally speaking, LSTM has shown great success for sequential data, leveraging long range dependencies and preserving the temporal order of the sequence (Cho et al., 2014; Graves, 2013). However, LSTM requires intensive computational resources both for training and inference due to its sequential nature.

As an alternative, recent NLP technologies such as machine translation (Gehring et al., 2017) and language modeling (Dauphin et al., 2016) have investigated other models such as convolutional neural networks (CNN). Convolutional-based architectures utilize hierarchical rather than sequential relationships between the input elements. Tem-poral Convolutional Networks (TCN) is a generic family of architectures that has been developed to alleviate the problem of training deep sequential models and is shown to provide significant improvement over LSTMs across different benchmarks (Bai et al., 2018; Lea et al., 2017). TCNs integrate *causal* convolutions where output at a certain time is convolved only with elements from earlier times in the previous layers. In addition, it is possible to train deep sequential models in parallel and use lower computational requirements rendering TCNs scalable to larger datasets.

---

[*]The work was done while the author was an intern at AWS, Amazon AI.

[1]With the exception of religious scripture and educational books for children, which are always diacritized.

In this study, we evaluate the application of the TCN architecture as described in (Bai et al., 2018) but devised as a character-level sequence model for diacritization since character-based models generalize better to unseen data compared to word-based models for most languages. Because diacritization is dependent on both past and future states, we further apply a variant of TCNs, namely *Acausal* Temporal Convolutional Network (A-TCN) (Lea et al., 2017), allowing the model to learn from previous as well as future context. We evaluate the architectures on three very different languages: Arabic, Vietnamese, and Yoruba.

To the best of our knowledge, TCN and A-TCN have not been investigated before for diacritization. In this paper, we show that A-TCN outperforms TCN while yielding comparable performance to BiLSTM but with the added advantage of being more efficient (speed and less computational footprint, i.e. decreased need for significant computational resources).

## 2 Related Work

A fair number of studies have been developed for the task of diacritization for different languages that include diacritics (Yarowsky, 1994; De Pauw et al., 2007; Scannell, 2011; Alqahtani et al., 2016, 2019). Feature engineering and classical machine learning algorithms such as Hidden Markov Models, Maximum Entropy Models, and Finite State Transducer were the dominant approaches (Nelken and Shieber, 2005; Zitouni et al., 2006; Elshafei et al., 2006). However, recent studies show significant improvement using deep neural networks (Belinkov and Glass, 2015; Pham et al., 2017; Orife, 2018). While these deep models achieve state-of-the-art performance, they mainly rely on the use of recurrent architectures such as BiLSTM, which are relatively inefficient.

Pham et al. (2017) view the task of diacritization for Vietnamese as a machine transduction problem from undiacritized to diacritized text at the word level. Orife (2018) addresses the problem on Yoruba in a similar way and compares soft- and self- attention sequence-to-sequence performance on the word level empirically showing that self-attention significantly outperforms BiLSTM.

Náplava et al. (2018) uses BiLSTM models with residual connections to train deeper models at the character level in several languages, including Vietnamese. For inference, they use beam search coupled with a language model to select among the possible diacritic patterns from the output. Zalmout and Habash (2017) develops a morphological disambiguation model to determine Arabic morphological features including diacritization. They train the model using BiLSTM while leveraging a LSTM-based language model as well as other morphological features to rank and score the output analyses.

## 3 Approach

We evaluate the performance of convolutional sequence models, TCN and A-TCN, in the context of diacritization and compare that to the recurrent sequential models, LSTM and BiLSTM. The task is formulated as a sequence classification such that we predict a diacritic for each character in the input.[2]

### 3.1 Convolutional Neural Network (TCN)

As TCN is a generic family of models, multiple architectures have been successfully developed (Dauphin et al., 2016; Kalchbrenner et al., 2016; Oord et al., 2016; Gehring et al., 2017; Lea et al., 2017; Bai et al., 2018). We choose the TCN architecture described in (Bai et al., 2018) [BAI18] as it integrates best techniques from the previous architectures while maintaining simplicity.

BAI18's TCN model satisfies the two main characteristics that allows sequential modeling: 1. Using 1-D fully convolutional network (Long et al., 2015) to ensure that the length of each layer is the same as the input by zero padding; 2. Using causal convolutions, which convolve output at time $t$ with elements from time $t-1$ and earlier in the previous layers. As Bai et al. (2018) indicated, developing a TCN architecture that only considers these two characteristics restricts how far the model can utilize previous information. Thus, to ensure learning from longer effective history, BAI18 further integrate dilated convolutions (Yu and Koltun, 2015), enabling an exponentially large receptive field correlated with the depth of the network. To enable deeper learning, they integrate a residual block (He et al., 2016), in which each block includes two layers of dilated causal convolution as fully illustrated in (Bai et al., 2018).

---

[2]We have also investigated BiLSTM-CRF, the model training time has been greatly increased without substantial improvement in terms of accuracy.

### 3.2 Acausal Convolutional Neural Network (A-TCN)

TCN is beneficial for applications that restrict information flow from the past, such as language modeling (Bai et al., 2018; Dauphin et al., 2016); however, this is not sufficient for diacritization. Incorporating future and past context has been previously integrated in different versions of TCN (Lea et al., 2017; Gehring et al., 2017) but it either did not enhance the performance of their tasks or it was evaluated on convolutional sequence-to-sequence models rather than sequential models as in the case of our study.

To incorporate both future and past context, we relax the causality constraint by integrating acausal convolution rather than causal convolution, hence A-TCN, as illustrated in Figure 1. A-TCN is a tweaked variant of TCN such that the model convolves information from $x_{t-d}$ to $x_{t+d}$ (previous and following states) instead of $x_{t-d}$ to $x_t$ (previous states) only.[3] In our implementation, we use layer normalization (Ba et al., 2016) rather than weight normalization. The extent in which context is considered is influenced by the number of layers as well as the kernel size. Each character is constructed from itself and $k-1$ characters, where *(k: kernel size)*. Additionally, as we go deeper, the model further incorporates different additional $k-1$ characters, skipping $d-1$ characters, where *(d: dilation factor)*, to incorporate the character at the $d^{th}$ position from both sides (see Figure 1).

## 4 Experimental Setup

**Dataset:** For Arabic, we use the Arabic Treebank (ATB) dataset: parts 1, 2, and 3 and follow the same data division as (Diab et al., 2013).

---
[3]Lea et al. (2017) investigated acausal convolution in their architecture but reported insignificant improvement over causal convolution in their problem space.

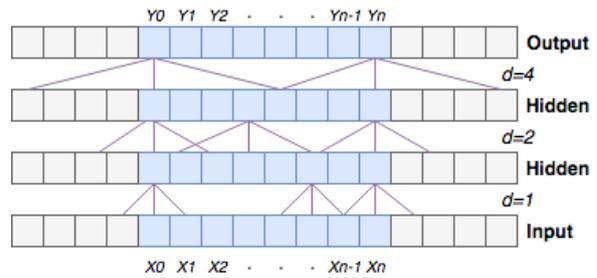

Figure 1: A dilated acausal convolution with 3 as a filter size and dilation factors equals to 1,2, and 4. White slots represent zero padding while colored slots represent characters. The remaining components of A-TCN are fully explained in (Bai et al., 2018).

| Data | Arabic | Vietnamese | Yoruba |
|------|--------|------------|--------|
| Train | 502,938 | 800,022 | 800,771 |
| Test | 63,168 | 786,236 | 44,598 |
| Dev | 63,126 | 408,093 | 44,314 |
| OOV | 7.3% | 1.3% | 2.6% |

Table 1: Number of word tokens as well as Out-Of-Vocabulary (OOV) rate. OOV rate indicates the percentage of undiacritized words in the test set not observed during training.

We use datasets provided by Orife (2018) and Jakub Nplava (2018) for Yoruba and Vietnamese, respectively. Most languages, especially those that are severely impacted by diacritics, rarely have such amounts of diacritized datasets. Thus, we sample a moderate size subset of the training data for Vietnamese, roughly 3.7% to train the models. In the process, we remove from the training set all sentences that have at least one word of more than 10 characters,[4] or that do not have at least one diacritic, or that contain more than 70 words. Table 1 illustrates the dataset statistics.

To augment the dataset without requiring additional annotated dataset, we segment each sentence into space tokenized units;[5] each unit is further segmented into its characters and passed through the model along with a specific number of previous and future words. We add the special word boundary "<w>" between words with a window size of 10 words before and after the target word (21 words in total).[6]

**Parameter Setting:** The character embedding vectors are of dimension size 45, randomly ini-tialized with a uniform distribution between [-0.1,0.1]. We use Adam optimization (Kingma and Ba, 2014) with 0.001 learning rate. For LSTM and BiLSTM, we use 3 layers and 250 hidden units. For TCN and A-TCN, we use 3 layers, 500 hidden units, and a kernel size of 5. Hidden units are initialized randomly using Xavier (Glorot and Bengio, 2010) with a magnitude of 3. For regularization, the dropout is set to 0.5 (Srivastava et al., 2014). We increase the dilation factor in TCN exponentially with the depth of the network.

---
[4]Vietnamese is characterized by having short word length.
[5]We did not investigate different types of tokenization. Relying on white space is simple and avoids additional preprocessing steps.
[6]We empirically tuned the window size on dev data.

## 5 Results

| System | DER | WER | OOV |
|---|---|---|---|
| **Arabic** | | | |
| Pasha et al. (2014) | - | 12.3% | 29.8% |
| Zalmout and Habash (2017) | - | 8.3% | **20.2%** |
| LSTM | 19.2% | 51.9% | 86.6% |
| TCN | 17.5% | 47.6% | 87.2% |
| BiLSTM | **2.8%** | **8.2%** | 33.6% |
| A-TCN | 3.0% | 10.2% | 36.3% |
| **Vietnamese** | | | |
| Náplava et al. (2018) | 11.2% | 44.5% | - |
| LSTM | 13.3% | 39.5% | 33.1% |
| TCN | 11.1% | 32.9% | 32.4% |
| BiLSTM | 2.6% | 7.8% | **15.3%** |
| A-TCN | **2.5%** | **7.7%** | **15.3%** |
| **Yoruba** | | | |
| Orife (2018) | - | 4.6% | - |
| LSTM | 13.4% | 37.2% | 84.9% |
| TCN | 12.7% | 35.5% | 83.8% |
| BiLSTM | **3.6%** | **12.1%** | **69.3%** |
| A-TCN | 3.8% | 12.6% | 70.2% |

Table 2: Models' performance on all words and OOV words per language. For Vietnamese, Náplava et al. (2018) reports 2.45% for WER on a much larger dataset (∼25M words), which is significantly better than our model.

We use two standard accuracy measures for evaluation: Diacritic Error Rate (DER) and Word Error Rate (WER). Table 2 shows the performance of the considered architectures across the sample languages.

**Importance of modeling past and future context:** A-TCN yields significant improvement over TCN, which indicates the importance of considering future and previous context for diacritization. Similarly BiLSTM significantly outperforms LSTM across all three languages. Overall, architectures that allow only unidirectional information flow provide lower performance than those that utilize context from both directions.

**Recurrent vs. convolutional architectures:** In line with prior work, across all three languages, TCN outperforms LSTM in the unidirectional architectures. A-TCN yields comparable results to BiLSTM except in the case of Arabic where the WER drops by ∼2%.

**OOV rate performance** To evaluate the model's robustness beyond observed training data, we specifically compare their WER performance on Out-Of-Vocabulary (OOV) words. BiLSTM has better ability to generalize on unseen data compared to A-TCN in Arabic and Yoruba, whereas both architectures are comparable in Vietnamese.

**Qualitative Analysis:** For Arabic, we further examine the impact of both BiLSTM and A-TCN on syntactic diacritics, aka inflectional diacritics which reflect syntactic case and mood primarily. These inflectional diacritics typically occur word finally in broken plurals and singular nouns as diacritics indicating syntactic case. Verbal mood is also marked word finally on verbs. We approximate the inflectional diacritics by computing the percentage of incorrectly predicted diacritics for the last characters of each word. BiLSTM yields a better performance (5.1% WER) compared to A-TCN (5.9% WER).

In addition, we randomly choose 20 sentences (605 words) to examine their categorical errors. We found similar errors in both architectures which include passive and active voice (e.g. نَشَرَت / na$arat[7] "spread" and نُشِرَت / nu$irat "been spread"), inflectional diacritics (e.g. وُصُولَ / wuSuwla and وُصُولُ / wuSuwlu, both mean "arrival" but have different syntactic positions), and named entities.[8] Thus, incorporating information from longer history in TCN architectures such as A-TCN compared to recurrent models such BiLSTM did not enhance learning inflectional diacritics.

We analyzed the confusion matrix and found that both architectures BiLSTM and A-TCN exhibit similar trends in the types of generated errors. In Arabic, the top two diacritics that are in-correctly diacritized by both A-TCN and BiLSTM (normalized by their frequencies) are $N$ and $\sim N$ which represent indefiniteness with and without doubling the consonants. The least diacritics that are incorrectly diacritized are no diacritics, sukun which represents the absence of diacritics, and two short vowels $i$ and $a$. This is consistent with the frequency of diacritics in Arabic except that in the diacritic $u$ which is also frequent but falls in the middle range of errors; this diacritic usually represents passive voice in Arabic which relies on syntactic relations between words. The same behavior can be found in Vietnamese and Yoruba where top and least erroneous diacritics are shared between both architectures.

---

[7]We adopt Buckwalter Transliteration encoding into Latin script for rendering Arabic text http://www.qamus.org/transliteration.htm.

[8]Diacritics in named entities are usually not consistent even among native speakers.

| Lang | BiLSTM | A-TCN | Difference | |
|------|--------|-------|------|------|
| | | | Time | Efficiency |
| AR | 376.85 | 132.55 | -64.8% | +284% |
| VI | 4187.81 | 1542.34 | -63.2% | +271% |
| YO | 461.19 | 138.45 | -70.0% | +334% |

Table 3: Inference time in seconds for each architecture across languages. *Efficiency* is defined in terms of text diacritization rate (amount of text diacritized per minute).

**Efficiency Comparison:** We provide a comparative analysis of models' training and inference runtime (Table 3). During training time, A-TCN yields similar efficiency gains across all three languages with comparable performance in terms of accuracy. The convergence criteria was set as at least 1% improvement in accuracy from the previous epoch. During inference time, A-TCN was 2.7∼3.3X faster in diacritizing text while providing comparable accuracy. This leads to 271∼334% improvement in terms of text diacritization rate in the amount of text diacritized per minute. This supports our overall observation that A-TCN is a solid alternative to BiLSTM for this problem space due to its efficient performance especially for industrial settings where time is a crucial factor. All experiments were carried out on a single Tesla P100 GPU.

For Arabic, for instance, BiLSTM took ∼19 hours to converge while A-TCN took ∼14 hours. The DER were 2.8% and 3.0% respectively. Thus, A-TCN was 24% faster than BiLSTM while being marginally lower by 0.2% in terms of DER.

**Comparison to Prior Work:** Table 2 shows the performance of previous models trained on the same data. For Arabic, both A-TCN and BiLSTM provide significantly better performance than MADAMIRA (Pasha et al., 2014), which is a morphological disambiguation tool for Arabic. The performance of Zalmout and Habash (2017)'s model falls in between BiLSTM and A-TCN. As opposed to our character-based models, both previous models use other morphological features along with a language model to rank all possible diacritic choices. We believe that this additional semantic and morphological information help their models perform better on OOV words.

For Vietnamese, when we re-train Náplava et al. (2018) model on the same sample discussed in Section 4, both A-TCN and BiLSTM provide significantly better results. Náplava et al. (2018) also use BiLSTM but with different parameter settings and different dataset preparation.[9] For Yoruba, both character-based architectures provide lower performance than Orife (2018)'s model. However, Orife (2018) uses seq2seq modeling which generate diacritized sentences that are not of the same length as the input and can generate words not present in the original sentence (hallucinations). This is unpleasant behaviour for diacritization especially if used in text-to-speech applications.

## 6 Conclusion

In this study, we show that character-based convolutional architectures for diacritization yield comparable performance to both word and character based RNN ones for multiple languages, albeit at a significantly lower computational cost. Moreover, character based modeling yields better performance overall for the diacritization task.

In all cases, A-TCN performs much better than TCN, with a reduction of up to 40% in error rate, which means that using future information is crucial in diacritization. All in all, the decision whether to use A-TCN or BiLSTM is a trade-off between accuracy and efficiency. A-TCN provides efficient solutions with comparable accuracy and comes in more handy for applications that need predicting diacritics on the fly such as text-to-speech applications and simultaneous machine translation. A-TCN was 2.7∼3.3X faster than BiLSTM at inference time (271%∼334% improvement in the amount of text diacritized per minute), crucially at comparable accuracies, making it a very attractive solution for industrial applications in particular.

---

[9]We augment the dataset as explained in Section 4.